%% file: main.tex
\PassOptionsToPackage{x11names}{xcolor}
\documentclass[10pt,twocolumn,letterpaper]{article}

\usepackage[pagenumbers]{cvpr} %

\input{preamble}

\definecolor{cvprblue}{rgb}{0.21,0.49,0.74}
\usepackage[pagebackref,breaklinks,colorlinks,citecolor=cvprblue]{hyperref}

\renewcommand{\thefootnote}{\fnsymbol{footnote}}

\title{Physically Plausible Full-Body Hand-Object Interaction Synthesis}

\makeatletter
\def\and{%
  \end{tabular}%
  \hskip 0.0em \@plus.17fil%
  \begin{tabular}[t]{c}}%
\makeatother
\author{%
  \hspace{0mm} Jona Braun$^{1}$ \and Sammy Christen$^{1}$ \and Muhammed Kocabas$^{1,2}$ \hspace{0mm} \and
   Emre Aksan$^{1 \dagger}$ \and Otmar Hilliges$^1$
  \vspace{1mm}
  \and
  $^1$ETH Zurich \quad $^2$Max Planck Institute for Intelligent Systems, Tübingen \vspace{-1mm} \\ \vspace{-3mm}
  \makebox[0cm]{\tt\footnotesize \{jona.braun, sammy.christen, muhammed.kocabas, otmar.hilliges\}@inf.ethz.ch aksan@google.com} \vspace{-2mm}
}

\begin{document}

\input{sec/figures/teaser}
\maketitle
\cftsetindents{section}{1em}{1.8em}
\cftsetindents{subsection}{1em}{2.4em}
\etocdepthtag.toc{mtchapter}
\input{sec/00_abstract}

\def\thefootnote{$\dagger$}\footnotetext{The project was completed after joining Google.}

\input{sec/01_introduction}

\input{sec/02_related_work}

\input{sec/03_method}

\input{sec/04_experiments}

\input{sec/05_conclusion}

{
    \small
    \bibliographystyle{ieeenat_fullname}
    \bibliography{main}
}

\appendix
\input{sec_supp/X_overview}
\etocdepthtag.toc{mtappendix}
\etocsettagdepth{mtchapter}{none}
\etocsettagdepth{mtappendix}{subsection}
{
  \hypersetup{
    linkcolor = black
  }
  \tableofcontents
}
\input{sec_supp/X_method_details}

\input{sec_supp/X_reward_function}

\input{sec_supp/X_implementation_details}

\input{sec_supp/X_additional_exp}

\input{sec_supp/X_future_work}

\end{document}

%% file: preamble.tex
\usepackage{epsfig}
\usepackage{graphicx}
\usepackage{amsmath}
\usepackage{amssymb}
\usepackage{multirow}
\usepackage{booktabs}
\usepackage{color, colortbl}
\usepackage[dvipsnames,x11names]{xcolor}
\usepackage{bm}
\usepackage{etoc}
\usepackage{float}
\usepackage{textcomp}
\usepackage{tocloft}

\newcommand{\figref}[1]{Fig.~\ref{#1}}
\newcommand{\secref}[1]{Section~\ref{#1}}

\newcommand{\eqnref}[1]{Eq.~\eqref{#1}}
\newcommand{\tabref}[1]{Tab.~\ref{#1}}

\newcommand{\supmat}{supp.~material\xspace}

\newcommand{\greencheck}{{\color{Green4} \checkmark}}
\newcommand{\redcross}{{\color{red}$\times$}}
\definecolor{Gray}{gray}{0.9}

\newcommand{\ttilde}{\raisebox{-0.7ex}{\~{ }}}

\newcommand{\policyvec}{\boldsymbol{\pi}}
\newcommand{\V}[1]{\mathbf{#1}} %
\newcommand{\R}{\rm I\!R}

\newcommand{\norm}[1]{\left\lVert#1\right\rVert}

%% file: sec/figures/teaser.tex
\twocolumn[{%
\maketitle
\begin{figure}[H]
\hsize=\textwidth 
\begin{center}
      \includegraphics[width=2.0\linewidth]{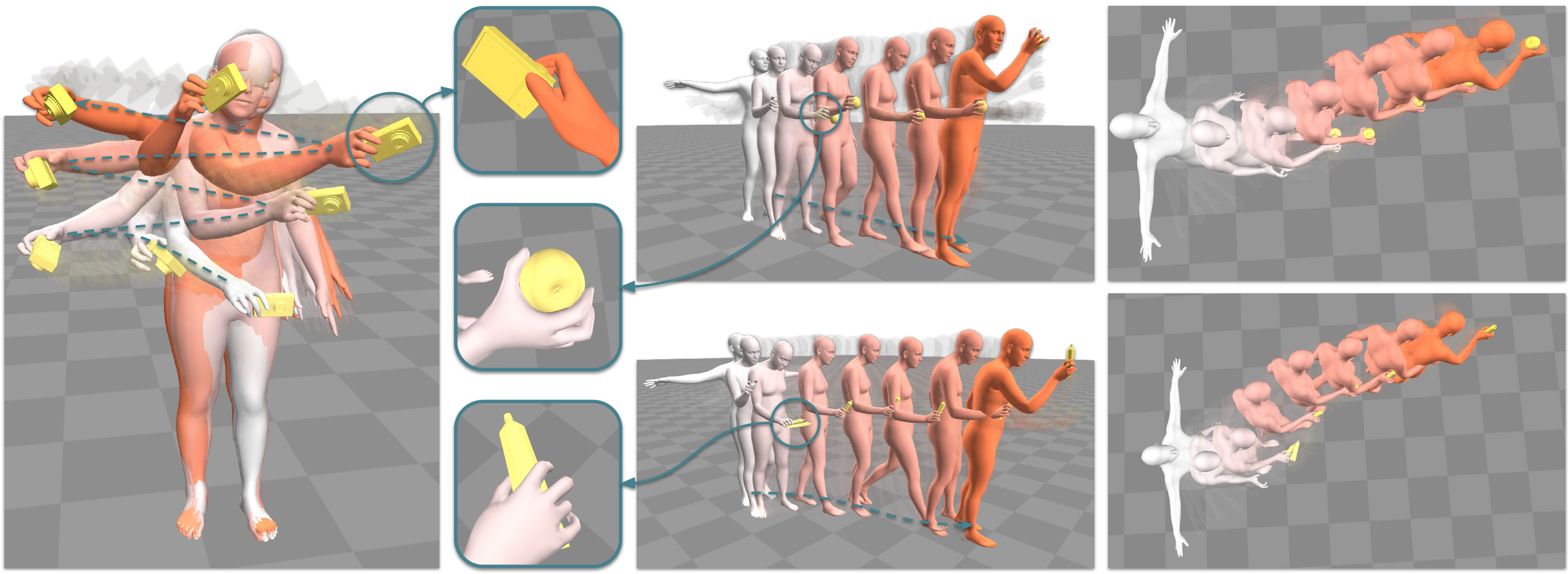}
\end{center}
    \vspace{-0.2cm}
   \caption{Our method generates physically plausible full-body hand-object interaction sequences. We can synthesize sequences with unseen objects while following a flexibly definable wrist trajectory (left). We can also generate motions of approaching an object, grasping it, then walking to a different location while lifting the object (middle, right). The target trajectories are indicated by the dashed lines.}
\label{fig:teaser}
\end{figure}
}]

%% file: sec/00_abstract.tex
\begin{abstract}

We propose a physics-based method for synthesizing dexterous hand-object interactions in a full-body setting. While recent advancements have addressed specific facets of human-object interactions, a comprehensive physics-based approach remains a challenge. Existing methods often focus on isolated segments of the interaction process and rely on data-driven techniques that may result in artifacts. In contrast, our proposed method embraces reinforcement learning (RL) and physics simulation to mitigate the limitations of data-driven approaches. Through a hierarchical framework, we first learn skill priors for both body and hand movements in a decoupled setting. The generic skill priors learn to decode a latent skill embedding into the motion of the underlying part. A high-level policy then controls hand-object interactions in these pretrained latent spaces, guided by task objectives of grasping and 3D target trajectory following. It is trained using a novel reward function that combines an adversarial style term with a task reward, encouraging natural motions while fulfilling the task incentives. Our method successfully accomplishes the complete interaction task, from approaching an object to grasping and subsequent manipulation. We compare our approach against kinematics-based baselines and show that it leads to more physically plausible motions.
Video and code are available at \url{https://eth-ait.github.io/phys-fullbody-grasp/}.

\end{abstract}

%% file: sec/01_introduction.tex
\section{Introduction}

Human-object interactions are at the core of our interactions with the physical world. Humans naturally interact with their environment through actions like approaching objects, grasping and manipulating them. 
The ability to simulate and comprehend these interactions has far-reaching implications in human-computer interaction, robotics, animation and AR/VR. 

While recent data-driven works have shown promising results in modeling certain aspects of human-object interactions, a comprehensive, physics-based full-body grasping approach covering the entire interaction process remains a challenge. 
Synthesizing dexterous grasps with full-body control is inherently challenging as it requires learning various tasks, namely balancing and moving the body naturally towards the objects, precise finger control, and performing a natural-looking and physically plausible grasp.

Recent works focus on distinct stages of human-object interaction, spanning from the initial approaching phase until grasping \cite{GOAL, SAGA} to the lifting of objects \cite{IMoS}, or even synthesizing the entire sequence \cite{li2023task}. Yet these efforts are primarily data-driven where the intricate physical constraints must be learned from the training data. Such purely data-driven settings can lead to artifacts and unrealistic behaviors due to the inherent limitations of training data such as foot-skating and interpenetration. In contrast, another line of research, physics-based human motion synthesis leverages physics simulation via reinforcement learning (RL) to mitigate limitations of data-driven paradigms. Existing works have either investigated human-object interactions at a larger scale \cite{Hassan:SIG:2023, luo2022embodied} or focused on dexterous hand grasping in an isolated manner \cite{D-Grasp, chen2022pregrasp}.

In this paper, we propose the first physics-based method to generate full-body human-object interactions for the entire task of approaching, dexterous grasping and manipulation of objects. By leveraging a physics simulation and reinforcement learning, our method synthesizes natural motions and mitigates physical artifacts, while ensuring that object motions emerge from forces applied by a humanoid agent. 

Our method adopts a hierarchical framework, where we first train low-level skill priors and then use these skill priors to learn full-body object interactions. At the core of our approach lies the decoupling of coarse body movement from fine-grained finger control. Specifically, we train separate general-purpose skill priors for the body and hand, decoding latent samples into body and hand movements. This approach ensures that small finger movements are not neglected in a unified training setup. We follow the adversarial training approach to learn these skill priors \cite{ASE}.

To enable full-body object interactions, we build a high-level policy for hand-object interactions that operates in the skill latent spaces. The outputs from this policy are translated into low-level control actions for the physics simulation. The high-level policy can be considered as planning module, leading the entire synthesis process. To guide the training of our high level policy, we propose a novel reward function that combines an adversarial reward to encourage natural motions with a reward to achieve stable grasps. 
To facilitate the training, we introduce a technique to explicitly condition the policy on 3D target trajectories for the root and wrist positions. This enables the policy to adapt to various scenarios and trajectories during inference.

In this work, we introduce a comprehensive, physics-based approach for the task of full-body grasp synthesis. Our method successfully accomplishes the complete interaction task, from approaching (unseen) objects to grasping and subsequent manipulation. We compare our method against the state-of-the-art techniques and present better performance, particularly in physics-based metrics, than the baselines. We further demonstrate the ability to follow diverse and unseen trajectories during inference, showcasing the flexibility and applicability of our method. Our main contributions are as follows:
\begin{itemize}
    \item A method to generate full-body, dexterous grasping interactions. To the best of our knowledge, this is the first physics-based approach to accomplishing the entire task.
    \item We propose a two-stage training scheme that decouples dexterous grasping from full-body motion during pretraining and uses joint training during finetuning.
    \item We compare our method against recent data-driven methods and show that our method produces more physically plausible results.
\end{itemize}

%% file: sec/02_related_work.tex
\section{Related Work}

We categorize related research into physics-based character control and motion synthesis. \tabref{tab:related_work} provides an overview of the most related works and ours. 

\input{sec/tables/tab_related}

\subsection{Physics Based Character Control}

Recent research \cite{TRPO, GAE, GAIL, PPO, catch_and_carry, DeepMimic, DreCon, MCP, residual_force, AMP, ASE} focuses on using deep reinforcement learning for physics based character control. \cite{catch_and_carry} train a humanoid to catch a tossed ball out of the air and then carry it to a target location. \cite{DeepMimic} show that incentivizing a policy to follow reference motions through the reward function can generate robust and natural behaviors. In follow-up work, AMP \cite{AMP} combine adverserial training to imitate reference motions with a task-specific reward. In ASE \cite{ASE}, AMP is scaled to train generalizable skill priors from large motion capture datasets. A high level policy is then trained on the skill prior to fulfill a task objective. In \cite{PADL}, this framework is extended to language conditioned inputs. In contrast to our work, these approaches do not consider finegrained dexterous grasping.

\subsection{Motion Reconstruction and Synthesis}
\paragraph{Kinematic based}
The synthesis of human body motion is a well-researched problem in computer vision \cite{Aksan_2019_ICCV, aksan2020motiontransformer, ghosh2017learning,holden2015tog, holden2016tog}.
Recent work has considered the synthesis of human-scene interaction \cite{hassan2021samp, cao2020long, starke2020local,zhang2022couch,wang2021scene,lee2023locomotionactionmanipulation, huang2023diffusionbased}, such as moving a box or sitting on a couch.
Contrary to our work, these methods do not consider finegrained hand-object interactions. FLEX \cite{tendulkar2023flex} jointly optimizes a hand and body pose prior to achieve diverse full-body grasping. Methods that use CVAEs to generate approaching motions for full-body grasps have been proposed \cite{GOAL,SAGA}. However, the generated motions only model the approaching phase and not the object manipulation phase. On the other hand, a recent work models the object manipulation phase conditioned on language commands \cite{IMoS}. In contrast to these works, we model the full interaction that includes the approaching and manipulation of an object, similar to \cite{li2023task}, but employ a physics simulation to increase the physical plausibility of outputs.

\paragraph{Physics-based}
Recent efforts have been made in leveraging physics simulations for various tasks such as pose estimation  ~\cite{SimPoe, shimada2020physcap, shimada2021neural, PoseTriplet, Dynamics_regulated}, human motion synthesis ~\cite{xie2021iccv}, and human-object interaction \cite{D-Grasp, chao2021learning,luo2022embodied,Hassan:SIG:2023}. Artifacts in pose reconstruction pipelines can for example be corrected by a physics-based policy \cite{SimPoe, PoseTriplet, Dynamics_regulated,luo2022embodied}. \cite{SimPoe} use off-the-shelf pose estimation as input to a pretrained imitation learning policy to obtain physically-plausible body motion. \cite{luo2022embodied} extend this by considering indoor scene interactions. \cite{zhang2023vid2player3d} learn physically plausible tennis skills from broadcast videos. Most closely related to ours, \cite{Hassan:SIG:2023} employ latent skill embeddings from large mocap data \cite{ASE} and train a high level policy to learn coarse object interaction, such as sitting on a couch or carrying a box.
On the other hand, recent works focus on the generation of hand-object interaction sequences in an isolated manner \cite{D-Grasp, mandikal2020graff, garciahernando2020iros, rajeswaran2017rss, qin2021dexmv}. Approaches often learn dexterous manipulation from full human demonstrations collected via teleoperation \cite{rajeswaran2017rss} or from videos \cite{garciahernando2020iros, qin2021dexmv}. \cite{mandikal2020graff} propose a reward function that incentivizes policies to grasp in the affordance region of objects. \cite{D-Grasp} propose a reinforcement learning based solution to generate diverse hand-object interactions from sparse reference inputs.
However, these approaches either model hand-object interactions but omit the body motion, or focus on the body motion and neglect fine-grained hand-object interactions. In contrast, we generate motions that model full-body hand-object interactions.

%% file: sec/tables/tab_related.tex
\begin{table}
\centering
\begin{center}
\resizebox{0.45\textwidth}{!}{%
\begin{tabular}{l|cccc}
   \toprule
    & \multirow{2}{*}{Full Body}  & \multirow{2}{*}{Physics}  & Whole  & Dexterous \\
     Method      &   &   & Interaction & Grasping  \\
    \midrule
    ManipNet \cite{Manipnet}    & \redcross    & \redcross                & \greencheck & \greencheck    \\
    D-Grasp \cite{D-Grasp}    & \redcross    & \greencheck                & \greencheck & \greencheck    \\
    GOAL \cite{GOAL}    & \greencheck    & \redcross                 & \redcross & \greencheck    \\
    SAGA \cite{SAGA}      & \greencheck  & \redcross                & \redcross & \greencheck \\
    IMOS \cite{IMoS} & \greencheck & \redcross             & \redcross   & \greencheck   \\
    Li \etal \cite{li2023task}  & \greencheck & \redcross             & \greencheck   & \greencheck   \\
    Hassan \etal \cite{Hassan:SIG:2023}       & \greencheck & \greencheck               & \greencheck   & \redcross   \\
    \textbf{Ours} &  \greencheck & \greencheck         & \greencheck & \greencheck  \\
   \bottomrule
\end{tabular}
}
\end{center}
\vspace{-0.2cm}
\caption{\textbf{Method Comparison.} We put our method into context with kinematics-based and physics-based approaches. Our method is the first to achieve physics-based full-body dexterous grasping.}
\label{tab:related_work}
\end{table}

%% file: sec/03_method.tex
\section{Task Setting}
\label{sec:task_setting}

We model the task of full-body human-object interaction as an RL-problem and leverage a physics simulation for training.
We are given an object with global pose $\textbf{T}_o \in \R^{6}$ and a human model $\bm{\Theta}=(\mathbf{t}_b, \bm{\theta}_b, \bm{\theta}_h)$, containing the global translation $\textbf{t}_b \in \R^{3}$, the body joint rotations $\bm{\theta}_b \in \R^{21x6}$ and finger joint rotations $\bm{\theta}_h \in \R^{16x6}$. We use the continuous 6D representation for rotations \cite{zhou2019continuity}. We base the model on the SMPL-X \cite{SMPLX} human body model but exclude eyeballs and jaw. Furthermore, we are provided with a hand pose reference $\bm{\Psi}$ and a target trajectory $\bm{\xi}$. 
The hand-object pose reference captures a single frame of a static hand grasp \cite{D-Grasp} and is defined as $\bm{\Psi}=(\overline{\bm{\theta}}_h, \overline{\textbf{t}}_h^0, \overline{\textbf{T}}_o)$, where $ \overline{\textbf{T}}_o$ is the reference object pose, $\overline{\bm{\theta}}_h$ and $\overline{\V{t}}_h^0$ indicate the target wrist joint rotations and translation, respectively. The target trajectory contains $n$ global target body and wrist 3D positions $\bm{\xi}=[\mathbf{{t}}_{b}^{i},\mathbf{{t}}_{h}^{i}]_{i=1}^{n}$. The goal of the task is to generate an output sequence of human and object poses
$[\bm{\Theta}_{}^{t}, \mathbf{T}_{o}^{t}]_{t=1}^{T}$ over horizon $T$. We split the task in two phases; in the first phase, the human character has to walk to the surface with the object and reach a grasp on the object. In the second phase, it has to manipulate the object by consecutively reaching the targets in the trajectory $\bm{\xi}$.

\input{sec/figures/method_overview}

\subsection{Simulation Environment}
\label{sec:simulation}
In the following we describe the environment of the physics simulation in which we train our human character.
We generate a controllable human body model following \cite{SimPoe}. It contains 57 DoF actuators for the body joints and 48 DoF actuators for the fingers, totaling 105 DoF. The root of the human (i.e., global 6DoF translation and orientation) is not actuated and changes according to the control of the other body joints. To reduce the computational complexity we approximate the collision geometries of the rigid body meshes with the exception of the ankles and feet. We focus on right-hand grasping and thus omit the left hand's fingers. We decimate all the object meshes to increase simulation speed. 
We use proportional derivative (PD) controllers to compute the torques  $\boldsymbol{\tau}$ to actuate the joints: 
\begin{align}
\begin{split}
    \boldsymbol{\tau} &= k_p \circ ({\bm{\hat{\theta}}} - \bm{\theta}) - k_d  \dot{\bm{\theta}}\\
    \bm{\hat{\theta}} &= \bm{\theta}_{\text{ref}} + k_s\mathbf{a}
\label{eq:pd_target}
\end{split}
\end{align}
where $\bm{\hat{\theta}}$ indicate the target joint rotations, $\bm{\theta}$ the current joint rotations, $ \dot{\bm{\theta}}$ the velocity and $k_p, k_d, k_s$ the gains. The target comprises the reference pose $\bm{\theta}_{\text{ref}}$ and  residual actions $\mathbf{a}$, which are predicted by our policies. The reference pose $\bm{\theta}_{\text{ref}}$ equals the current pose $\bm{\theta}_h$ for the finger control and the center between the joint limits for the body joint control.
The state space of the simulation is given by $\mathbf{s}=(\bm{\Theta}, \dot{\bm{\Theta}}, \textbf{T}_o, \dot{\textbf{T}}_o, \mathbf{f})$, which contains the human pose and velocity information, the object pose and velocity, and the net contact force $ \mathbf{f} \in \R^{39x1}$ acting on the human body joints, the object, and the table surface. See \supmat for more details about the simulation environment.

\subsection{Reinforcement Learning}
We follow \cite{sutton1998introduction} and model RL as a Markov Decision Process (MDP) defined by a 6-Tuple $\mathcal{M} = (\mathcal{S}, \mathcal{A}, \mathcal{T}, \mathcal{R}, \mu, \gamma)$, where $\mathcal{S}$ is the state and $\mathcal{A}$ the action space. The deterministic transition function $\mathcal{T}: \mathcal{S} \times \mathcal{A} \mapsto \mathcal{S}$ maps from a state-action pair to the next state and the reward function $\mathcal{R}: \mathcal{S} \times \mathcal{A} \mapsto \mathbb{R}$ maps to a scalar value. The first state $\mathbf{s}_0$ is determined by the initial state distribution $\mu: \Delta(\mathcal{S}) \mapsto \mathcal{S}$. Finally, $\gamma \in [0,1]$ defines the discount factor of future rewards.
We define a parametric policy $\policyvec: \mathcal{S} \mapsto \Delta(\mathcal{A})$ that maps to a distribution over actions given a state. We aim to optimize the policy such that it maximises the expected discounted reward $\mathbb{E}_{\mathbf{a}_t \sim \pi_{\boldsymbol{\theta}}(\cdot\mid \mathbf{s}_t), \mathbf{s}_0 \sim \mu}\left[\sum_{t=0}^{T-1} \gamma^t \mathcal{R}(\mathbf{s}_t, \mathbf{a}_t)\right]$ where $\mathbf{s}_{t+1} = \mathcal{T}(\mathbf{s}_t,\mathbf{a}_t)$ and $T$ is the horizon.

\section{Full-Body Grasp Motion Synthesis}
\label{sec:full_body_grasp}

Our framework is inspired by ASE \cite{ASE} and depicted in \figref{fig:method}. Therefore, we leverage a hierarchical framework. First, we train low-level priors that represents diverse motion skills from motion capture data. Thereafter, we train a high-level policy, dubbed hand-object interaction policy, that predicts actions in the latent spaces of the priors to achieve a high-level objective. In our setting, the objective is to approach the object, grasp it and move it according to a specified wrist and root trajectory. We now first explain how we train physics-based body and hand priors and then describe our hand-object interaction policy training.

\subsection{Pre-Training of Body and Hand Priors}

In our approach, we decouple the training of the body prior and the hand prior. Crucially, this prevents mode collapse and allows learning coarse body movements and finegrained finger control. Each prior is represented by a policy $\pi(\mathbf{a} \mid \bm{\phi}({\mathbf{s})}, \mathbf{z})$, which is conditioned on features extracted from the physics simulation's state $\bm{\phi}({\mathbf{s})}$ and a latent skill vector $\mathbf{z} \sim p(\mathbf{z})$. 
We combine a motion imitation objective and an unsupervised skill discovery objective \cite{ASE} to train these priors. The motion imitation objective incentives the policy to perform motions that are similar as depicted in the reference motion. It is optimized by training a discriminator to differentiate between motions sampled from the reference motion capture data and motions generated by the humanoid character. The skill discovery objective promotes the policy to learn a meaningful latent skill space which allows a high-level policy to reuse the learned skills. Thus, the reward function is defined as follows:
\begin{align}
\begin{split}
    r = &-\log(1-D(\bm{\phi}({\mathbf{s})},\bm{\phi}(\mathbf{s'})) \\
    &+ \beta \log q(\mathbf{z}_t\mid \bm{\phi}(\mathbf{s}),\bm{\phi}(\mathbf{s'})),
    \label{eq:discr}
\end{split}
\end{align}
where $D$ indicates the discriminator and $q$ is an encoder trained with the objective to recover the latent skill vector $\mathbf{z}$ from a tuple of features $(\bm{\phi}(\mathbf{s}),\bm{\phi}(\mathbf{s'}))$ from the simulation state $\mathbf{s}$ and the consecutive state $\mathbf{s'}$.

\paragraph{Hand Prior}
The hand prior is a policy $\pi_h(\mathbf{a}_h \mid \bm{\phi}_h (\mathbf{s}),\mathbf{z}_h)$ that controls the wrist and the finger joints via the actions $\mathbf{a}_h$. It is conditioned on the latent skill vector $\mathbf{z}_h$ and the right-hand features $\bm{\phi}_h (\mathbf{s})=(\bm{{\theta}}_h, \bm{\dot{\theta}}_h, \mathbf{x}_h)$, where $\bm{{\theta}}_h \in \R^{16x6}$ and $\bm{\dot{\theta}}_h \in \R^{16x6}$ indicate the local hand joint rotations (except for the global wrist joint orientation) and their angular velocities, and $\mathbf{x}_h \in \R^{16x3}$ are the wrist-relative 3D finger joint positions. 
To train the hand prior, we detach the hand from the body and fix its global position in space. For training, we use the reward function in \eqnref{eq:discr} with hand-state tuples $(\bm{\phi}_h(\mathbf{s}),\bm{\phi}_h(\mathbf{s'}))$.

\paragraph{Body Prior}
We extend our body prior setting to a goal-conditioned approach by explicitly considering the 3D target positions of the root $\mathbf{t}_{b}$ and the wrist $\mathbf{t}_{h}$ as conditional variables. Our body-prior policy $\bm{\pi}_b(\mathbf{a}_b \mid \bm{\phi}_b(\mathbf{s}), \mathbf{z}_b, \mathbf{t}_{b}, \mathbf{t}_{h})$ controls all body joints except the hands. Similarly, the body encoder $q_b$ is conditioned on the target positions such that $q_b(\mathbf{z}_b \mid \bm{\phi}_b(\mathbf{s}), \bm{\phi}_b(\mathbf{s'}), \mathbf{t}_{b}, \mathbf{t}_{w})$. We further leverage this additional information by introducing an auxiliary reward on the target positions during high-level policy training (see \secref{sec:high_level_policy}). The benefits of including the root and wrist targets in the conditional variables are twofold. First, both the policy and the encoder gain spatial awareness, reducing ambiguity and yielding better planning. Second, this formulation allows us to control generated motion at inference time, e.g., walking to a target root position or moving the right wrist to a target position. 

The body-state features are defined as $\bm{\phi}_b(\mathbf{s})=(\bm{\theta}_{b}, \dot{\bm{\theta}}_{b}, \mathbf{x}_b, \dot{\mathbf{x}}_b, \mathbf{h}_b, \dot{\mathbf{t}}_b)$. The terms $\bm{\theta}_{b}$ and $\dot{\bm{\theta}}_{b}$ indicate the root-relative body joint rotations and their velocities  (except for the global root joint orientation and velocity). $\mathbf{x}_b$ and $\dot{\mathbf{x}}_b$ are 3D joint positions and their velocities (excluding the root). $\mathbf{h}_b$ is the root's height (e.g., the value in z-direction according to our preprocessing) and $\dot{\mathbf{t}}_b$ is the root's linear velocity. All the features except the root height and root orientation are in the root-relative coordinate-frame. The body-state features for the discriminator are a subset of the policy features $\bm{\phi}_b(\mathbf{s})$, similar to \cite{ASE}. For training, we use the reward function in \eqnref{eq:discr} with body-state tuples, $(\bm{\phi}_b(\mathbf{s}),\bm{\phi}_b(\mathbf{s'}))$. See \supmat for more details.

\subsection{Training of Hand-Object Interaction Policy}
\label{sec:high_level_policy}
We leverage the body and hand prior to train a hand-object interaction policy. The policy $\bm{\pi}_{\text{ho}}(\mathbf{z}_b,\mathbf{z}_h, \mathbf{\tilde{t}}_{b},\mathbf{\tilde{t}}_{h} \mid \bm{\phi}_b(\mathbf{s}),\bm{\phi}_h(\mathbf{s}), \bm{\phi}_{\text{ho}}(\mathbf{s}, \bm{\Psi}, \bm{\xi}))$ is conditioned on both hand and body features $\bm{\phi}_b(\mathbf{s}),\bm{\phi}_h(\mathbf{s})$ and task-relevant features $\bm{\phi}_{\text{ho}}(\mathbf{s}, \bm{\Psi}, \bm{\xi})$ (see below). It predicts the latent vectors $\mathbf{z}_b$ and $\mathbf{z}_h$ of both the body and hand prior. These latent vectors are then passed to the policies which yield output actions that are applied to the human body model. Additionally, we predict position targets $\mathbf{\tilde{t}}_{b}$ and $\mathbf{\tilde{t}}_{h}$ for the root and wrist as a training scheme which we dub target guidance. 

\paragraph{Hand-Object Features}
The features $\bm{\phi}_{\text{ho}}(\mathbf{s})$ represent the task-relevant information that is required for grasping the object and following a target trajectory:
\begin{equation}
   \bm{\phi}_{\text{ho}}(\mathbf{s}, \bm{\Psi}, \bm{\xi}) = (\textbf{T}_o, \dot{\textbf{T}}_o, \mathbf{g}_x, \mathbf{g}_{\theta}, \mathbf{g}_c, \mathbf{g}_{\xi}, \mathbf{f}_h, \mathbf{x}_\text{tab}, \bm{\eta}). 
\end{equation}

\noindent The 6D root-relative object pose and its velocity are given by $\textbf{T}_o$ and $\dot{\textbf{T}}_o$. The terms $ \mathbf{g}_x, \mathbf{g}_{\theta},$ and $\mathbf{g}_c$ are features computed from the static hand pose reference $\bm{\Psi}$ (see \secref{sec:task_setting}) to measure the distance between the current hand pose and the target hand pose:
\begin{align}
    \mathbf{g}_x = \overline{\mathbf{x}}_h-\mathbf{x}_h \quad \mathbf{g}_{\theta} = \overline{\bm{\theta}}_h \ominus \bm{\theta}_h \quad \mathbf{g}_{c} = (\overline{\bm{c}}_h, \overline{\bm{c}}_h \ominus \bm{c}_h)
    \label{eq:goal_feats}
\end{align}
\noindent The distance between the 3D joint positions of the reference pose $\overline{\mathbf{x}}_h$ and the current pose $\mathbf{x}_h$ in root-relative frame is given by $\mathbf{g}_x$. The 6D rotational difference between the reference hand pose $\overline{\bm{\theta}}_h$ and the current hand pose $\bm{\theta}_h$ is defined by $\mathbf{g}_{\theta}$. Similarly, $\mathbf{g}_{c}$ is a tuple containing the contact targets and the distance between the target and the current contacts. It is a vector with binary values indicating whether a target contact is achieved or not.
The target 3D joint positions $\overline{\mathbf{x}}_h$ and the target contacts $\overline{\mathbf{c}}_h$ are computed from the hand pose reference $\bm{\Psi}$. Note that contacts in our context are on a per-joint basis. 

Similarly, to guide the human character along a given trajectory, it is provided with the distance to the next waypoints on the trajectory $\mathbf{g}_{\xi}$:
\begin{equation}
     \mathbf{g}_{\xi} = (\mathbf{{t}}_{b}^{i}-\mathbf{{t}}_{b},\mathbf{{t}}_{h}^{i}-\mathbf{{t}}_{h}),
     \label{eq:goal_traj}
\end{equation}
\noindent where $\mathbf{{t}}_{b}^{i}$ and $\mathbf{{t}}_{h}^{i}$ are the next root and wrist targets to achieve. Once a target has been reached, the next one is sampled from the trajectory $\bm{\xi}$.

Lastly, $\mathbf{f}_h$ is the vector describing the net forces acting on the hand joints, the object, and the table surface (see \secref{sec:simulation}). The term $\mathbf{x}_\text{tab}$ is the distance between the 3D wrist joint and the table. The phase variable $\bm{\eta} \in [0, 1]$ depicts the progress of the task. We provide more details on the hand-object state features $\bm{\phi}_{\text{ho}}(\mathbf{s})$, in \supmat.

\paragraph{Hand-Object Reward Function}
To guide the policy to grasp the object and follow the trajectory $\bm{\xi}$, we define the following hand-object reward function:
\begin{align}
    r_{\text{HO}} = w_T \, r_T(\mathbf{s},\mathbf{a}) + w_S \, r_S(\mathbf{s}),
\end{align}
where $r_T$ and $r_S$ indicate the task and style reward with weights $w_T$ and $w_S$, respectively.

The task reward $r_T$ incentivizes the policy to achieve a stable grasp on the object and follow the target trajectory:
\begin{align}
    r_T =  r_{x} + r_{\theta} + r_c + r_{\xi}+ r_{\text{reg}}, \label{eq:task_reward}
\end{align}
where the terms $r_{x}$, $r_{\theta}$, $r_c$, and $r_{\xi}$ are position, orientation, contact and trajectory rewards, respectively. These rewards are computed by taking the norm of the distance features introduced in \eqnref{eq:goal_feats} and \eqnref{eq:goal_traj}.
Lastly, $r_{\text{reg}}$ indicates a regularization reward on the predicted actions. Details on the reward function are provided in the \supmat.

We introduce a style reward $r_S$ to achieve more plausible and natural motions. It extends the discriminator-based style reward of \cite{ASE} for the hand. Specifically, we use the discriminator predictions for the hand and body such that 
\begin{align}
    \begin{split}
    r_S = &-\log(1-D_b(\bm{\phi}_b(\mathbf{s}),\bm{\phi}_b(\mathbf{s}'))) \\ &-\log(1-D_h(\bm{\phi}_h(\mathbf{s}),\bm{\phi}_h(\mathbf{s'}))).
    \end{split}
\end{align}
\paragraph{Target Guidance}
We introduce target guidance to allow the policy to be robust and flexibly follow the given targets $\bm{\xi}=[\mathbf{{t}}_{b}^{i},\mathbf{{t}}_{h}^{i}]_{t=1}^{n}$ for the root and wrist joints.
During training, we alternate the target trajectory $\bm{\xi}$ between the ground-truth $(\mathbf{\overline{t}}_{b},\mathbf{\overline{t}}_{h})$ and the predicted targets $(\mathbf{\tilde{t}}_{b},\mathbf{\tilde{t}}_{h})$ and regularize the training with an auxiliary objective:
\begin{equation}
\begin{aligned}
    \mathcal{L}_{\xi} = ||{\overline{\bm{\xi}} - \bm{\tilde{\xi}}} ||_2^2,
\end{aligned}
\end{equation}
where $\bm{\tilde{\xi}}$ is the target prediction and $\overline{\bm{\xi}}$ is the ground truth trajectory. The loss measures the Euclidean distance between the two terms. Note that target guidance is applied only after after the object has been grasped.

\subsection{Implementation Details}
We follow the actor-critic framework \cite{sutton1998introduction} and implement our skill priors with 4-layer MLP networks using [1024, 1024, 512, 512] units and ReLu activations after every layer. In the actor network, we use a Gaussian output model with a constant variance and predict only the mean.
The discriminators and encoders share the first 3 linear layers [1024, 1024, 512] with separate final layers. 
The high-level hand-object policy $\bm{\pi}_{\text{ho}}$ is implemented with a 3-layer MLP and a Gaussian output model with constant variance where the final layer predicts the mean. For training, we use the Adam optimizer \cite{kingma2015iclr} with a learning rate of 2e-5 and a discount factor $\gamma$ of $0.99$. We implement our method in PyTorch \cite{paszke2019pytorch}. We use Isaac Gym \cite{makoviychuk2021isaac} as physics simulation. It runs at 120Hz while the policies are sampled at 30Hz. Further details can be found in the \supmat.

%% file: sec/figures/method_overview.tex
\begin{figure*}[t]
\begin{center}
   \includegraphics[width=0.9\linewidth]{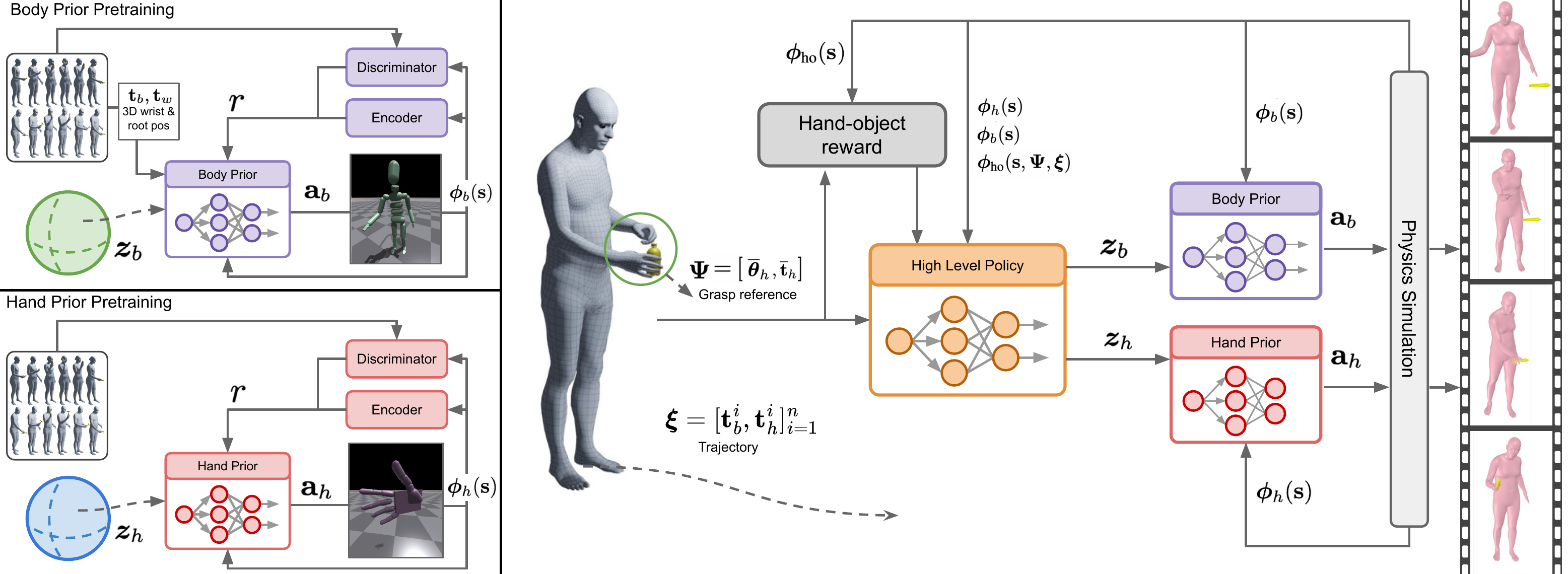}
\end{center}
   \vspace{-4mm}
   \caption{\textbf{Method Overview.} Given a hand-object pose reference $\bm{\Psi}$ and a root and wrist target trajectory $\bm{\xi}$, our method synthesizes a human approaching, grasping an unseen object and following a trajectory with it. The training procedure involves learning skill priors, which we decouple into a body prior and a hand prior. We train the priors from a motion capture dataset using adversarial training.
   The right-hand prior policy $\pi_h(\mathbf{a}_h \mid \bm{\phi}_h (\mathbf{s}),\mathbf{z}_h)$ predicts the right-hand target joint angles $\bm{a}_h$. The body prior policy $\pi_b(\mathbf{a}_b \mid \bm{\phi}_b(\mathbf{s}), \mathbf{z}_b, \mathbf{t}_{b}, \mathbf{t}_{w})$ predicts target joint angles of the body $\bm{a}_b$. PD-controllers compute the necessary torques to drive the joints to the predicted target angles in the physics simulation.  Instead of directly conditioning the policies on the physics simulation state $\bm{s}$, we define feature extraction functions $\phi(\cdot)$ for each policy. A latent vector $\mathbf{z}$ is used to represent the skill space. The body prior is additionally conditioned on a root and wrist 3D target positions $\bm{t}_b$ and $\bm{t}_h$. 
   To train a hand-object interaction policy $\bm{\pi}_{\text{ho}}(\bm{z}_b,\bm{z}_h, \mathbf{t}_{b}, \mathbf{t}_{h} \mid \phi_h(\mathbf{s}),\phi_h(\mathbf{s}), \phi_{\text{ho}}(\bm{s}, \bm{\Psi}, \bm{\xi}))$, we predict the latent vectors $\bm{z}_b$ and $\bm{z}_h$ of the body and hand prior, respectively. We predict position targets $\mathbf{\tilde{t}}_{b}$ and $\mathbf{\tilde{t}}_{h}$ for the root and wrist as an auxiliary objective. We define a task reward function based on the physics simulation state $\mathbf{s}$, the static hand pose reference $\bm{\Psi}$, and the trajectory $\bm{\xi}$.}
\label{fig:method}
\end{figure*}

%% file: sec/04_experiments.tex
\section{Experiments}
We first describe the data and experimental details in Sections \ref{sec:data} and \ref{sec:exp_details}. \secref{sec:evaluation} presents our main evaluations, consisting of quantitative and qualitative comparisons against the baselines. Lastly, in \secref{sec:ablation}, we provide an ablation to highlight the contributions of our method.

\input{sec/tables/tab_evaluation}

\subsection{Data}
\label{sec:data}
We train and evaluate our model using the GRAB dataset \cite{GRAB} where we follow the right-handed grasp setting as in the prior works \cite{GOAL, SAGA, li2023task}. We combine the object test-split from GOAL \cite{GOAL} and the subject test-split from IMoS \cite{IMoS}. Hence, our training set contains all sequences from subjects S1-S9 and the object-split of GOAL. We then evaluate on both the GOAL and IMoS test sets.

Our humanoid character in the physics simulation is based on the neutral SMPL-X model. Hence, we convert the subject-specific GRAB reference motions to the neutral model. 
This preprocessing involves aligning the feet with the ground and the object with the hand. We provide more details on the preprocessing in the \supmat.

\input{sec/figures/qualitative_results_motion}

\subsection{Experimental Details}
\label{sec:exp_details}
During training of the hand-object interaction policy, we initialize the character at a random frame of the approaching phase sampled from a GRAB reference clip. The object and table are initialized according to the hand pose reference. We use a two-stage training procedure. First, we fix the object to its surface, such that the character can learn to approach and initiate a stable grasp on the object without the risk of moving or dropping the object. In the second stage, the object is non-stationary such that the policy learns to lift and follow the trajectory. To avoid overfiting, we add random noise to the hand pose reference, the target trajectory, the initial object position and rotation around the yaw axis. The noise applied to the object position is also added to the table position to prevent interpenetration of the object. 

The one-to-one correspondence between the neutral SMPL-X model and our humanoid in the physics simulation enables a direct conversion between the two. Hence, we are able to run evaluations in the SMPL-X parameter space (except for the grasping success and the TTR metric, see \secref{sec:metrics}) and compare our method against the kinematics-based approaches. 
At evaluation time, the humanoid agent is always initialized in T-pose and its root is set to the root of the initial test frame.
Finally, we apply Gaussian smoothing to the output motion as a post-processing step. We find that the smoothing operation marginally improves the performance. Our model's performance without the smoothing operation is reported in \supmat.

\subsubsection{Baselines}
Our method is capable of modeling the entire task of approaching an object, grasping and manipulating it. In contrast, the relevant baselines focus on a particular phase, e.g., GOAL \cite{GOAL} generates motions for the \emph{approaching} phase while IMoS \cite{IMoS} tackles object \emph{manipulation} after grasping. 
Hence, we compare our method against one baseline from each phase for a fair comparison. 
Though related, \cite{li2023task} is a very recent submission with no code publicly available. 

We evaluate the baselines using the publicly available source code, pre-trained models and following the proposed evaluation protocols. Please note that there are differences between the settings of our method and IMoS. We model the entire task with a focus on single-handed object manipulation by providing an explicit control on the target trajectories. On the other hand, IMoS introduces language based control for two-handed object manipulation. Despite these differences, we deem a comparison justified since the physics-based metrics we report are invariant to the setting.
 \\

\subsubsection{Metrics}
\label{sec:metrics}
We use the metrics proposed in prior works \cite{SAGA, GOAL, D-Grasp, jiang2021graspTTA}. The formal definitions are provided in \supmat. \\
\textbf{Grasp Success Rate:} We consider a grasp a success when the object is held for at least $0.5$s in the physics simulation without dropping. For our model this includes approaching the object and lifting it from the table. 
We determine the success rate of the kinematics baselines using a static pose as a reference in physics simulation. The humanoid character and object are initialized with the last generated motion frame and maintain the grasp via PD-control \cite{D-Grasp, jiang2021graspTTA}. \\
\textbf{Ground Distance (GD):} We compute the distance between the average floating height (above ground) and the average vertical ground penetration depth, which are determined by the lowest SMPL-X vertex. \\
\textbf{Foot Skating (FS):} The percentage of foot skating frames. We consider a foot to be skating if the lowest SMPL-X vertex exceeds a threshold velocity \cite{GOAL}. \\
\textbf{Interpenetration:} We report the interpenetration volume (IV) of MANO vertices that penetrate the object mesh and the maximum interpenetration depth (ID). In the \emph{approaching} phase, we average the metric across the last five frames to be able to capture interpenetration before reaching the final grasp. For the \emph{manipulation} phase, we average over five evenly distributed frames. \\
\textbf{Trajectory Targets Reached (TTR):} The ratio of the targets reached over all the targets in the trajectory. If a target is not reached within a certain time window, it is considered a failure and the next target from the trajectory is sampled. This metric is only applicable to our method and in the \emph{manipulation} phase. \\
\textbf{Contact Ratio (CR):} The ratio of hand vertices that are within 5mm of the object mesh averaged over the sequence. \\

\subsection{Evaluation}
\label{sec:evaluation}
We provide a qualitative results of our method in \figref{fig:eval_qualitativ_motion} and a comparison against the baselines in \figref{fig:eval_qualitative_interpenetration}. Please see our supplementary video for more examples.

We compare our method with GOAL \cite{GOAL} in the approaching phase until grasping and with IMoS \cite{IMoS} in the manipulation phase after grasping. Note that while we evaluate each phase separately, our method always performs the full sequence. We report the results in \cref{tab:motion_synthesis} using the metrics outlined in \secref{sec:metrics}. We also provide the metrics for the ground truth (GT) as reference.

\noindent\textbf{Physical Plausibility}
Our method outperforms both baselines in all metrics, highlighting benefits of having a physics simulation in-the-loop. It leads to fewer artifacts as indicated by the hand-object interpenetration volume (IV) and depth (ID), foot skating (FS), and ground distance (GD). Baseline results often exhibit ground penetration, floating above ground, and hand-object collisions (see \figref{fig:eval_qualitative_interpenetration}). Notably, our method also displays better physics-based properties compared to the ground truth data, which we argue is due to noise in the motion capture and labeling. 
Note that as a consequence of the approximated collision geometry as rigid bodies in the physics simulation, our method can still exhibit small amounts of interpenetration after converting the simulation results to the SMPL-X parameter space. 

\input{sec/figures/qualitative_comparison}

\noindent\textbf{Contact Ratio}
To be in line with related work, we report the contact ratio (CR). We find that ours has a lower CR in the approaching phase than GOAL and a comparable CR with IMOS in the manipulation phase. However, we argue that this metric may not correlate with grasp quality due to the wide range of grasps. For example, grasps that mainly involve fingertips, such as a pinch grasp, lead to a lower CR. Furthermore, we observe that GOAL sometimes penetrates the object while approaching, yielding a high contact ratio despite the violation of physical constraints.

\noindent\textbf{Success Rate}
Our method consistently achieves higher grasp success rates compared to the baselines. %
Note that simulation-based metrics such as grasp success have been established in previous works \cite{jiang2021graspTTA, D-Grasp} and give an indication on grasp stability. However, it should to be interpreted with care when comparing physics and kinematic methods directly, since physics-based methods leverage a simulation, whereas kinematic-based methods do not. Small amounts of noise in contacts may already cause failure, because the PD-controller only maintains the input pose.
Lastly, we validate how successful our method can follow a given target trajectory (TTR). The results indicate that most targets of the unseen test trajectories can be reached. 

\noindent\textbf{Generalization}
Our method can generalize to unseen objects (GOAL test set). It has difficulties grasping large objects where the fingers need to be fully stretched such as the \textit{large cube} or \textit{piggybank}. While these objects are part of the training set, they influence the success rate on the S10 test set. Examples of failure cases are in \supmat.

\subsection{Ablations}
\label{sec:ablation}
We report ablation results in \tabref{tab:ablation}. We analyze the decoupling of the body prior from hand prior (\textit{decoupling}), the two-stage training (\textit{two-stage}) and the target guidance (\textit{t-guid.}). We train all policies on the entire training set and evaluate on the test set. We find that decoupling of the coarse body motion from the dexterous hand motion is a critical component. Training a full-body prior directly leads to mode collapse in the latent space and hence fails to learn the full-body grasping task. The two-stage training procedure also plays an important role in achieving better performance. It allows the hand-object policy to first focus on achieving a stable grasp and then learn to follow the target trajectory. Lastly, our target guidance technique further improves the performance due to the explicit conditioning on target positions and the auxiliary training objective.

\input{sec/tables/tab_ablation}

%% file: sec/tables/tab_evaluation.tex
\begin{table*}[t!]
  \centering
    \resizebox{0.9\textwidth}{!}{%
  \begin{tabular}{@{}lccccccc@{}}
    \toprule
    Method & Success ($\uparrow$) & GD [mm] ($\downarrow$) & FS [\%]($\downarrow$) & IV [$cm^3$] ($\downarrow$) & ID [$mm$] ($\downarrow$) & TTR $\uparrow$ & CR \\ \midrule\midrule
    \multicolumn{8}{c}{Approaching}\\\midrule
    Ground-truth & 0.29 & 5.9 & 7.7 & 1.66 & 4.4 & - & 0.111 \\
    GOAL \cite{GOAL} & 0.13 & 6.8  & 14.4 & 1.97 & 5.3 & - & 0.128 \\
    Ours & \textbf{0.79} & \textbf{2.1/0.0}* & \textbf{5.7} & \textbf{0.11} & \textbf{1.1} & - & 0.026 \\\midrule
    \multicolumn{8}{c}{Manipulation}\\\midrule
    Ground-truth (S10 test set) & 0.22 & 6.1 & 2.9 & 2.75 & 4.9 & - & 0.112 \\
    IMOS \cite{IMoS} (S10 test set) & 0.20 & 16.0  & 8.0 & 5.07 & 6.8 & - & 0.057 \\
    Ours (S10 test set) & \textbf{0.64} & \textbf{1.8/0.0}* & \textbf{0.9} & \textbf{0.22} & \textbf{2.7} & \textbf{0.65} & 0.053 \\ \midrule
    Ours (GOAL test set) & 0.79 & 1.9/0.0* & 1.2 & 0.18 & 2.9 & 0.85 & 0.055 \\
    \bottomrule
  \end{tabular}
  }
  \vspace{-1mm}
  \caption{\textbf{Evaluation.} We compare our method against the relevant baselines on approaching until grasping and manipulation after grasping. In both settings, we find that our method achieves better performance across all of the metrics. We also provide the metrics for the ground-truth motion capture data (GT) as reference. Notably, our method can correct artifacts present in motion capture data, such as ground penetration or floating. The success rates show that our method leads to most stable grasps in the physics simulation. * The ground distance (GD) in the SMPL-X space is not zero as a consequence of the rigid body approximation of the human in the physics simulation. This metric equates to 0.0 when evaluated directly in the physics engine.}
  \vspace{-3mm}
  \label{tab:motion_synthesis}
\end{table*}

%% file: sec/figures/qualitative_results_motion.tex
\begin{figure}[t]
\begin{center}
   \includegraphics[width=0.97\linewidth]{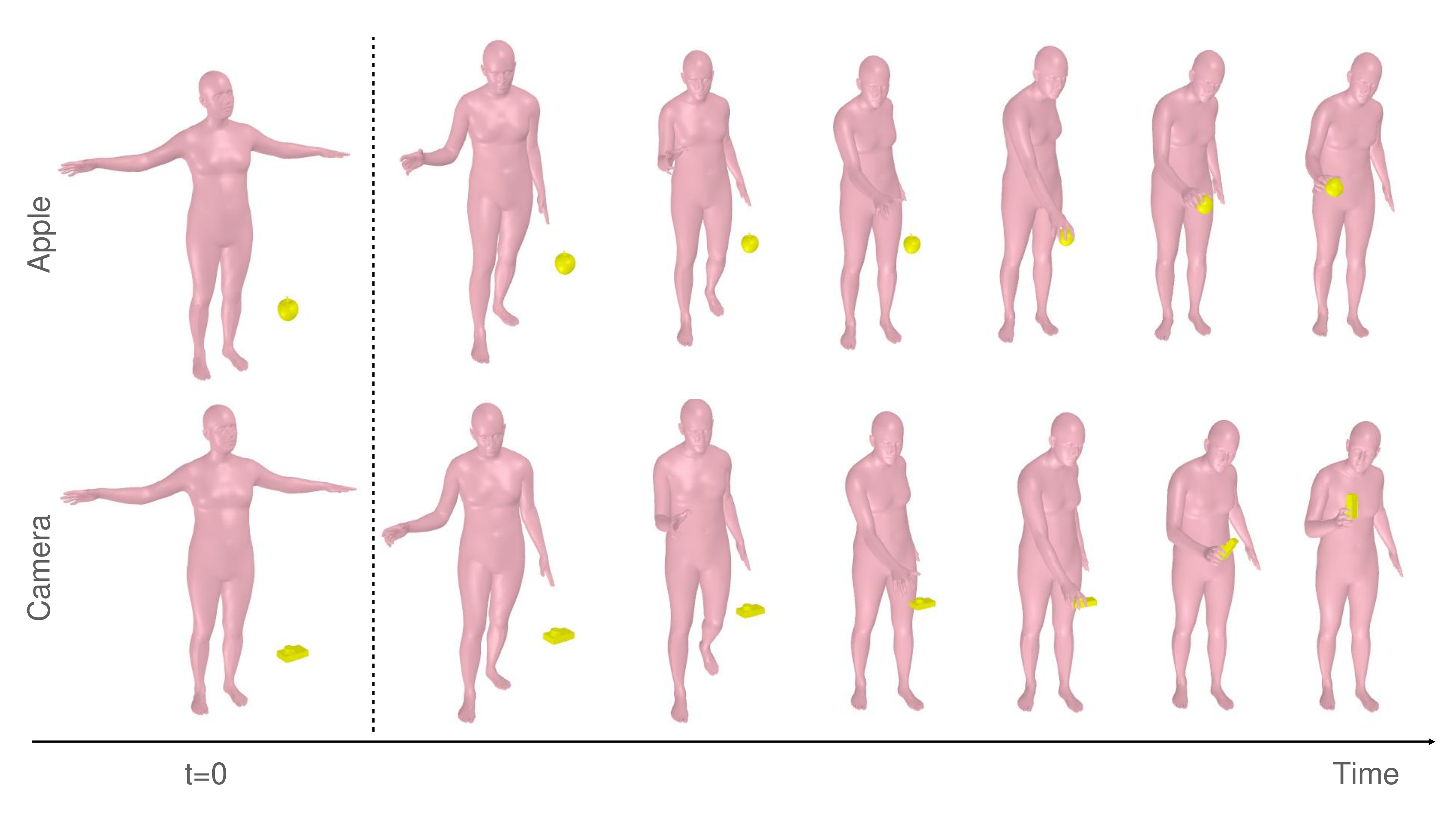}
\end{center}
\vspace{-7mm}
   \caption{\textbf{Qualitative Results.} Each row shows a motion sequence generated by our model for an unseen object from the test set. }
   \vspace{-3mm}
\label{fig:eval_qualitativ_motion}
\end{figure}

%% file: sec/figures/qualitative_comparison.tex
\begin{figure}[t]
\begin{center}
   \includegraphics[trim={0.5cm 3.5cm 21cm 0.1cm}, clip, width=0.92\linewidth]{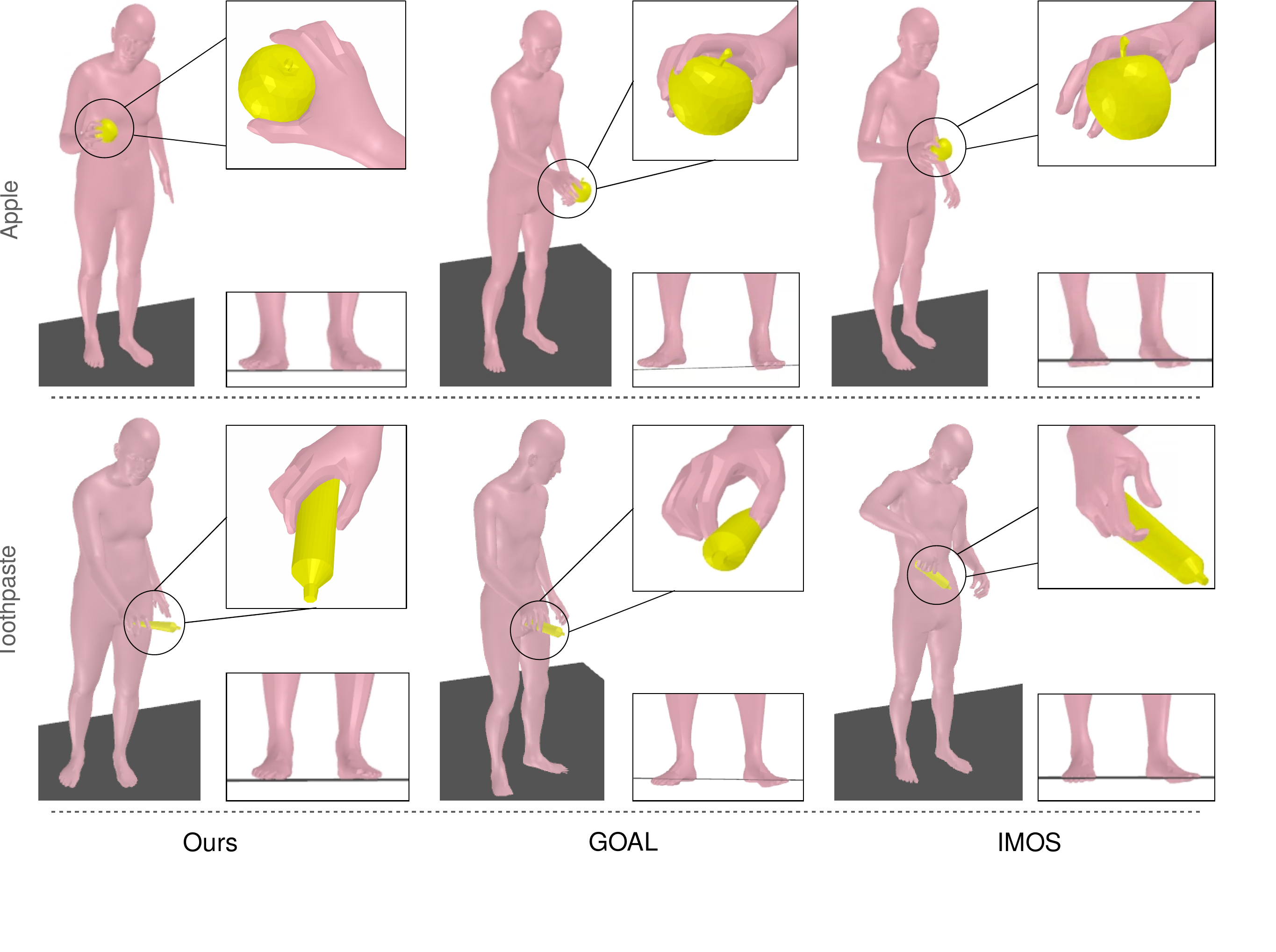}
\end{center}
\vspace{-10mm}
   \caption{\textbf{Qualitative Comparison.} Our physics-based method generates motions that exhibit less hand-object interpenetration and ground interpenetration than the kinematics baselines. }
   \vspace{-3mm}
\label{fig:eval_qualitative_interpenetration}
\end{figure}

%% file: sec/tables/tab_ablation.tex
\begin{table}
  \centering
  \resizebox{0.85\linewidth}{!}{%
  \begin{tabular}{@{}ccc|ccc@{}}
    \toprule
    \textit{decoupling} & \textit{two-stage} & \textit{t-guid.}  & Success $\uparrow$ & TTR $\uparrow$ \\ \midrule
    \redcross & \redcross & \redcross & 0.0 & 0.0 \\
    \greencheck & \redcross & \redcross & 0.55 & 0.56 \\
    \greencheck & \greencheck & \redcross& 0.77 & 0.79 \\
    \greencheck & \greencheck & \greencheck & \textbf{0.79} & \textbf{0.85} \\
    \bottomrule
  \end{tabular}
  }
  \vspace{-0.1cm}
  \caption{\textbf{Ablations.} We ablate the components of our method. The decoupling of priors is crucial to solve the task, while the two-stage training procedure and target guidance each contribute to higher success rates in grasping and trajectory following.}
  \vspace{-2mm}
  \label{tab:ablation}
\end{table}

%% file: sec/05_conclusion.tex
\section{Discussion and Conclusion}
\label{sec:conclusion}

We have introduced the first method to achieve physics-based full-body dexterous grasping. Our approach involves a hierarchical framework, beginning with the training of decoupled skill priors for body and hand control. These priors are then leveraged to develop a high-level policy to orchestrate the approaching, grasping and trajectory-guided manipulation phases. Notably, our method demonstrates a promising degree of physical plausibility in comparison to kinematics-based baselines. Our work also opens the door to potential future directions. For instance, there is potential in conditioning policies on language prompts, as shown in \cite{PADL, IMoS}, to guide the humanoid character. Moreover, our existing model relies on a single hand reference pose for guidance, a limitation that we hope could be addressed in future work. Lastly, while our current focus remains on single hand grasping, learning how to achieve physics-based bi-manual full-body grasping remains an open challenge. 

%% file: sec_supp/X_overview.tex
\clearpage
\setcounter{page}{1}
\maketitlesupplementary

We provide this \textbf{manuscript} and a \textbf{video} as supplementary material. The table of contents below contains the structure of this document. Our code and models will be made publicly available upon publication.

%% file: sec_supp/X_method_details.tex
\section{Method Details}

\subsection{Discriminator Observations}
\paragraph{Hand Prior} The hand-prior discriminator features  $\bm{\phi}_h^D(\mathbf{s}) = (\bm{{\theta}}_h, \bm{\dot{\theta}}_h, \mathbf{x}_h^D)$ are equal to the hand-prior state features $\bm{\phi}_h(\mathbf{s})$ with the exception that only wrist-relative 3D joint positions $\mathbf{x}_h^D$ of fingertips (instead of all joints) are used. This design choice is motivated by \cite{ASE}, which uses a pruned version of the full state for the discriminator.

\paragraph{Body Prior} 

The body-prior discriminator features are similar to the body-prior state features and defined as $\bm{\phi}_b^D(\mathbf{s}) = (\bm{\theta}_{b}^D, \dot{\bm{\theta}}_{b}^D, \mathbf{x}_b^D, \mathbf{h}_b, \dot{\mathbf{t}}_b)$.
The terms $\bm{\theta}_{b}^D$ and $\dot{\bm{\theta}}_{b}^D$ represent the local (parent-relative instead of root-`relative as in $\bm{\phi}_b(\mathbf{s})$) joint orientations and their angular velocities (except for the global root joint orientation and velocity). The root-relative 3D joint positions of key joints (left and right: elbow, wrist, knee, ankle, foot) are indicated by $\mathbf{x}_b^D$. The height of the root is defined by $\mathbf{h}_b$ and the linear velocity of the 3D root position is given by $\dot{\mathbf{t}}_b$.

\subsection{Body-Prior Reward Function}
Besides the discriminator and encoder rewards outlined in \eqnref{eq:discr} of the main paper, the body prior uses a trajectory reward $r_{\xi}^b$ and  a regularization reward $r_{\text{reg}}^b$:
\begin{align}
\begin{split}
    r_b =  r_{\xi}^b +  r_{\text{reg}}^b &-\log(1-D(\bm{\phi}_b({\mathbf{s})},\bm{\phi}_b(\mathbf{s'})) \\
    &+ \beta \log q(\mathbf{z}_b\mid \bm{\phi}_b(\mathbf{s}),\bm{\phi}_b(\mathbf{s'})).
\end{split}
\end{align}

\paragraph{Trajectory Reward}
Given a randomly sampled 3D target root position $\mathbf{t}_{b}^{i}$ and target wrist position $\mathbf{t}_{h}^{i}$, the trajectory reward for the body prior is computed as the distance to the current root position $\mathbf{t}_{b}$ and wrist position $\mathbf{t}_{h}$: 
\begin{align}
     r_{\xi}^b = \exp{(-\alpha_b \ \Delta{t}_{b})} + \exp{(-\alpha_h \ \Delta{t}_{h})}, \label{eq:traj_reward_body_prior}\\
    \Delta{t}_{b} = \biggl(\min\biggl(\norm{\mathbf{t}_{b}^{i} - \mathbf{t}_{b}}, \beta_b\biggr) - \beta_b\biggr), \label{eq:traj_reward_delta_body}\\
    \Delta{t}_{h} = \biggl(\min\biggl(\norm{\mathbf{t}_{h}^{i} - \mathbf{t}_{h}}, \beta_h\biggr) - \beta_h\biggr),\label{eq:traj_reward_delta_hand}
\end{align}
where the weights are defined by $\beta_b=0.2$, $\beta_h=0.005$, $\alpha_b=2.0$, and $\alpha_h=3.0$. 

\paragraph{Regularization Reward}
To prevent fast, unnatural movements we regularize the linear wrist velocity $\dot{\mathbf{t}}_h$:
\begin{align}
     r_\text{reg}^b  = 1- \min \bigl ( {\exp{(\norm{\mathbf{t}_b} - 0.8)}, 10}\bigr). %
\end{align}

\subsection{Hand-object State Features}

\paragraph{Contact Features}
We now explain in more detail the contact features $\mathbf{g}_{c}$ from \eqnref{eq:goal_feats} of the main paper:

\begin{equation}
    \mathbf{g}_{c} = (\overline{\bm{c}}_h, \overline{\bm{c}}_h \ominus \bm{c}_h).
\end{equation}
The first term $\overline{\bm{c}}_h \in \R^{16x1}$ is a binary target contact vector, which indicates which hand joints (16 in total) should be in contact with the object according to the hand pose reference $\Psi$. The second term is a distance vector with binary values showing whether a target contact is achieved or not:

\begin{equation}
    \overline{\bm{c}}_h \ominus \bm{c}_h = \mathbb{I}_{\overline{\bm{c}}_h = \bm{c}_h=1}.
\end{equation}
For each contact body in  $\bm{c}_h$, the vector is 0 unless a target contact $\overline{\bm{c}}_h$ is achieved, in which case it is 1.

\paragraph{Motion-Phase}
The term $\bm{\eta} \in [0, 1]$ indicates which phase of the task the human character is in. To this end, we define a set of six discrete states using the following heuristics:
\begin{enumerate}
    \item  The distance between the wrist and object is above 0.5m.
    \item  The distance between the wrist and object is below 0.5m, but above 0.2m.
    \item  The distance between the wrist and object is below 0.2m.
    \item  The hand is in contact with the object.
    \item  The object is lifted from the table.
    \item  The vertical distance between the initial object position and the current position is larger than 3cm.
\end{enumerate}
To encode these states into the phase variable, we simply quantize the interval and assign it to the states in increasing order (i.e., the first state is assigned 0.0, the second state 0.2, etc.).

%% file: sec_supp/X_reward_function.tex
\subsection{Task Reward Function}
The task reward $r_T$ of the hand-object interaction policy (see \cref{eq:task_reward} in the main paper) is a linear combination between the static grasp reward (\cref{sec:static_grasp_reward}), the trajectory reward (\cref{sec:trajectory_reward}), and a regularization reward (\cref{sec:regularization_reward}).

\subsubsection{Static Grasp Reward} \label{sec:static_grasp_reward}
The static grasp reward incentivizes the policy to grasp the object firmly such that it does not slip out of the hand. The reward is split into joint position reward  $r_{x}$, joint orientation reward $r_{\theta}$, and a contact reward $r_c$.

\paragraph{Position Reward}
The position reward promotes moving the wrist and finger joints (including the fingertips) to the 3D target joint positions given by the hand pose reference $\bm{\Psi}$. To make the 3D target joint positions invariant with respect to the object pose, we convert all joint positions into object-relative frame.
Given the current 3D target joint positions $\mathbf{\overline{x}}_h^j$ and the current 3D target joint positions $\mathbf{x}_h^j$ of each joint $j$, we compute:
\begin{align}
         &\Delta{\mathbf{x}}_{a:b} = \sum_{j=a}^b \biggl(\min\biggl(\norm{\mathbf{\overline{x}}_h^j - \mathbf{x}_h^j}, \beta_x\biggr) - \beta_x\biggr),\\
        &\begin{aligned}
            r_x = 0.5 \ &\exp{(-1.25 \ \Delta{\mathbf{x}}_{1:J})} + \exp{(-1.5 \ \Delta{\mathbf{x}}_{1:1})},
        \end{aligned}
\end{align}
where $J$ is the total number of joints, $\beta_x = 0.01$m is a constant and $j=1$ indicates the wrist joint.

\paragraph{Orientation Reward}
The orientation reward $r_{\theta}$ incentivizes the policy to move the wrist and finger joints into the target orientations given by the hand pose reference $\bm{\Psi}$. We make use of the geodesic norm to compute the reward.
Given the current joint rotation $\boldsymbol{q}_h^j$ and the target joint rotation $\boldsymbol{\overline{q}}_h^j$ as quaternion of each joint $j$ (which we convert from $\bm{\theta}_h$ and $\overline{\bm{\theta}}_h$), we compute:
\begin{align}
\begin{split}
        &\Delta{{q}}_h^j = \arccos(2 \  (\boldsymbol{q}_h^j \circ \boldsymbol{\overline{q}}_h^j)^2 - 1)\\
        &\Delta{{q}}_{h} = \min\biggl(\frac{1}{J} \sum_{j=1}^J \Delta{q}_h^j, \beta_{\theta}\biggr) - \beta_{\theta}\\
        &r_{\theta} = \exp{(- 2.5 \ \Delta{q}_h)}, %
\end{split}
\end{align}
where $\circ$ indicates quaternion multiplication, $J$ is the total number of joints, $\beta_{\theta}=0.1$rad is a constant and $j=1$ indicates the wrist joint.

\paragraph{Contact reward}
The contact reward $r_c$ comprises three components: the \textit{contact-mask} reward $r_{\text{c,\text{mask}}}$, the \textit{contact force} reward $r_{\text{c,\text{force}}}$, and the \textit{no-table-contact} reward $r_{\text{c,\text{tab}}}$:
\begin{align}
r_c = r_{\text{c,\text{mask}}} + r_{\text{c,\text{force}}} + r_{\text{c,\text{tab}}} 
\end{align}

The \textit{contact-mask} reward guides the hand parts towards reaching the target contacts extracted from the hand pose reference $\bm{\Psi}$:
\begin{align}
    &\begin{aligned}
        c_{\text{mask}} = \biggl(\frac{\overline{\bm{c}}_h^\top \bm{c}_h}{\overline{\bm{c}}_h^\top\overline{\bm{c}}_h}
        + \frac{\overline{\bm{c}}_h^\top \bm{c}_h^{t-1}}{\overline{\bm{c}}_h^\top\overline{\bm{c}}_h}\biggr) %
        \label{eq:contact_mask}
    \end{aligned} \\
    &r_\text{c,\text{mask}} = 1 - \exp(-1.6 \ c_{\text{mask}}).
\end{align}

The term $\frac{\overline{\bm{c}}_h^\top \bm{c}_h}{\overline{\bm{c}}_h^\top\overline{\bm{c}}_h}$ computes the ratio of number of bodies in contact with the object according to the hand pose reference. $\bm{c}_h^{t-1}$ is the binary contact vector from the previous physics simulation state. Hence, the second term in \eqnref{eq:contact_mask} promotes coherent contacts over time. An entry in $\bm{c}_h$ is 1 if the net contact force for that joint body is larger than zero.\\

The \textit{contact force} reward incentivizes the policy to apply enough force between the hand and the object to grasp it stably:
\begin{align}
    &\eta_\text{force} =  \min(\norm{\mathbf{f}_{h}}, \beta_c \ m_{o})\\ %
    &r_\text{c,\text{force}} = \exp(0.25 \ \eta_\text{force}) \ \mathbb{I}_{\overline{\bm{c}}_h = \bm{c}_h=1}, \\
\end{align}
where $m_{o}$ is the object's weight and $\beta_c=15$ is a constant. In essence, the term $\eta_\text{force}$ promotes forces being applied up to an empirically defined maximum net force. The reward is only added for joints that are supposed to be in contact according to the target contacts $\overline{\bm{c}}_h$, which is indicated by $\mathbb{I}_{\overline{\bm{c}}_h = \bm{c}_h=1}$.\\

The \textit{no-table-contact} reward promotes being in contact with the object while avoiding forces applied to the table:
\begin{align}
    r_\text{c,\text{tab}} =  \mathbb{I}_{\mathbf{f}_{h}>0} \mathbb{I}_{\mathbf{f}_{\text{tab}}=0},
\end{align}
where $\mathbb{I}_{\mathbf{f}_{h}>0}$ is an indicator for hand-object contact and $\mathbb{I}_{\mathbf{f}_{\text{tab}}=0}$ is an indicator that is 1 if there is no force applied to the table by neither the object nor the hand. Note that the reward is non-zero only if both conditions are true.

\subsubsection{Trajectory Reward}\label{sec:trajectory_reward}
Given the current 3D root position $\mathbf{t}_{b}$, the current root-relative wrist position $\mathbf{t}_{h}$, and the current $i$-th trajectory target positions ($\mathbf{t}_{b}^{i}$, $\mathbf{t}_{h}^{i}$), we compute the reward as described in \eqnref{eq:traj_reward_body_prior}, but with different weights and an additional component:
\begin{align}
    r_\xi = \exp{(-\alpha_b \ \Delta{t}_{b})} + \exp{(-\alpha_h \ \Delta{t}_{h})} + \alpha_{s} \ N_{\text{success}},
\end{align}
where $\beta_b=0.01, \beta_h=0.01, \alpha_b=1.25, \alpha_h=3.0, \alpha_s=0.008$. The last term is used to counterbalance a drop in the position reward as soon as a target is reached and a subsequent target is sampled, because this may make the policy not pursue any targets. This reward term increases with the number of achieved targets $N_{\text{success}}$.

\subsubsection{Regularization Reward}
\label{sec:regularization_reward}
The regularization reward $r_{\xi,\text{reg}}$ is defined as follows:
\begin{align}
    r_{\text{reg}} = \exp(-\norm{\mathbf{\dot{t}}_o}) + \exp(-\norm{\mathbf{\ddot{t}}_h}).
\end{align}
 We regularize the object's linear velocity $\dot{\textbf{t}}_o$ and the jerk of the hand $\mathbf{\ddot{t}}_h$ (computed with finite differences from $\mathbf{t}_h$).

%% file: sec_supp/X_implementation_details.tex
\section{Implementation Details}

\subsection{Simulation environment}
The physics simulation environment contains the humanoid, the object and a table. We model the table as a floating box and the object using its mesh. The provided meshes in GRAB have a high vertex count. In order to reduce the computational complexity of collision detection, we decimate all meshes. We compute the object weight based on the mesh volume and a constant density. We base our humanoid on the neutral SMPL-X \cite{SMPLX} human body model but exclude eyeballs and jaw. The skeleton of the humanoid is created by extracting the joint positions and kinematic tree of the SMPL-X body model. We add an actuator to each joint and limit the joints based on the distribution of the GRAB dataset \cite{GRAB}. Similar to \cite{SimPoe}, we create a rigid body mesh for every joint of the SMPL-X body model. The body meshes are built by assigning each vertex to the joint with the largest linear blend skinning weight and then computing a convex hull per joint. The weight of each body is computed using the volume of the mesh and a constant density. To simplify the computational complexity, we approximate the collision geometries of the rigid body meshes with boxes, cylinders, and capsules, with the exception of the ankles and feet. Since we focus on right-hand grasping, we remove the left hand's finger joints from the humanoid.

As Isaac Gym \cite{makoviychuk2021isaac} does not yet allow to determine the origin of the net contact force experienced by a rigid body, we disable certain collisions in order to retrieve useful contact observations. All collision between the humanoid and table are disabled. Moreover, all self-collisions between hand joints are disabled during the training of the hand-object interaction policy. However, self-collisions of the fingers are enabled during pre-training of the hand prior, which should prevent learning skills that cause self-penetration. 
\subsection{Preprocessing}
As our humanoid character in the physics simulation is based on the neutral SMPL-X model, we need to convert the subject specific GRAB data. 
We first align the feet with the ground by translating each frame of the motion by the distance of the lowest SMPL-X vertex to the ground, i.e., we either lift or lower the character. To align the object with the hand, we translate the object and table by the distance between the thumb joints of the subject-specific and the neutral characters' motions. We determine the hand pose reference using a heuristic, where we choose the frame within a time-window after the initial hand-object contact with the highest number of hand-object contacts. To add variety to training, we add multiple hand pose references close to the chosen frame in time. Finally, we optimize the hand poses of the references using ContactOpt \cite{grady2021contactopt}. To generate target trajectories, we extract a set of wrist and root position targets that are $1/15$s apart from the motion capture reference motions, starting from the initial frame of hand-object contact. In our experiments, we limit the reference motions to a length of 4s. Instead of using one single set of targets per trajectory during training, we shift a window over the motion clip, which yields multiple sets of targets. 

\subsection{Training Setup}

We use a single 80GB A100 to train the body and hand prior and a 24GB RTX 3090 TI NVIDIA GPU to train the hand-object interaction policy.
We simulate 8192 parallel environments when training the priors and 2048 parallel environments for the hand-object interaction policy. The policies are updated after sampling 32 steps in each environment, yielding batches of \ttilde262k and \ttilde65k samples for the priors and the hand-object interaction policy, respectively. We train the priors for 40k and the hand-object interaction policy for 190k epochs, which amounts to roughly 6 days and 7 days of training, respectively.

\input{sec/tables/tab_no_smoothing}
\input{sec/tables/tab_trajectory_per_object}

\section{Experimental Details}

\paragraph{Randomization}
We randomly sample hand pose references and target trajectories during training. To increase robustness, we add uniform noise of $[-30, 30]$mm to the hand pose references and $[-2, 2]$mm to the trajectory targets, respectively.

\subsection{Metric Details}
\textbf{Grasp Success Rate:} We consider an object grasp successful if the object does not drop to the ground or table within a time window of 0.5s. For the baselines, we directly initialize the sequences in the predicted grasping pose without a table and consider a grasp successful if the object does not drop to the ground within 0.5s. \\
\textbf{Ground Distance (GD):} Given the set of SMPL-X 3D vertices $\mathcal{V}_i$ per frame $i$, we extract the z-coordinate of the lowest vertex as $z_i = \min_{z}(\mathcal{V}_i)$. We compute the metric as follows: 
\begin{align}
    \text{GD} = \frac{\sum_i z_i \mathbb{I}_{z_i > 0}}  { \sum_i \mathbb{I}_{z_i > 0}} - \frac{\sum_i z_i \mathbb{I}_{z_i < 0}}  {\sum_i \mathbb{I}_{z_i < 0}}.
\end{align}
It computes the distance between the average floating height and the average ground penetration depth. If $\sum_i \mathbb{I}_{z_i > 0} = 0$ or $\sum_i \mathbb{I}_{z_i < 0} = 0$, we use 0 for that term.\\
\textbf{Foot Skating (FS):} Given the set of SMPL-X 3D vertices $\mathcal{V}_i$ per frame $i$, we find the vertex with the lowest z-coordinate $\mathbf{v}_j=\text{argmin}_{z}(\mathcal{V}_i)$. The foot is considered skating if the horizontal velocity $\dot{\mathbf{v}}_j > 1$cm per frame as proposed in \cite{GOAL} (note that we ignore the z-component of $\dot{\mathbf{v}}_j$). We compute the percentage of frames that are foot skating over all frames $N_{\text{tot}}$:
\begin{align}
    \text{FS} = \frac{\sum_i^{N_{\text{tot}}} \mathbb{I}_{\dot{\mathbf{v}}_j > 1\text{cm}}} { N_{\text{tot}} }
\end{align}
\textbf{Interpenetration:} The interpenetration volume (IV) is computed as the average volume of vertices $\mathcal{V}$ penetrating the object mesh. The interpenetration depth (ID) is given by the maximum distance between penetrating vertices and the object surface. In the \emph{approaching} phase, we average the metric across the last five frames to capture interpenetration before reaching the final grasp. For the \emph{manipulation} phase, we average over five evenly distributed frames. \\
\textbf{Trajectory Targets Reached (TTR):}  Let $N_{\text{tot}}$ be the total count of all reached targets in the trajectory and $N_{\text{success}}$ the number of targets that were reached within a given time horizon of 0.2s, then $\text{TTR} = N_{\text{success}}/N_{\text{tot}}$. We consider a target reached if the wrist position is within 12cm of the target. \\ %
\textbf{Contact Ratio:} The ratio of SMPL-X vertices $\mathcal{V}_i$ per frame $i$ that are within 5mm of the object mesh, averaged over the whole sequence.

%% file: sec/tables/tab_no_smoothing.tex
\begin{table*}[t!]
  \centering
    \resizebox{0.8\textwidth}{!}{%
  \begin{tabular}{@{}lccccc@{}}
    \toprule
    Method & GD [mm] ($\downarrow$) & FS [\%]($\downarrow$) & IV [$cm^3$] ($\downarrow$) & ID [$mm$] ($\downarrow$) &  CR \\ \midrule\midrule
    \multicolumn{6}{c}{Approaching}\\\midrule
    Ours w/o smoothing  &  2.2 & \textbf{2.2} & 0.15 & 1.9 & 0.035 \\
    Ours  &  \textbf{2.1} & 5.7 & \textbf{0.11} & \textbf{1.1} & 0.026 \\\midrule
    \multicolumn{6}{c}{Manipulation}\\\midrule
    Ours w/o smoothing (S10 test set)  & {2.0} & {4.2} & \textbf{0.13} & \textbf{1.6} & 0.030 \\
    Ours (S10 test set) & \textbf{1.8} & \textbf{0.9} & 0.22 & 2.7 & 0.053 \\ \midrule
    Ours w/o smoothing (GOAL test set)  & 2.1 & 4.4 & \textbf{0.13} & \textbf{1.4} & 0.033 \\
    Ours (GOAL test set) & \textbf{1.9} & \textbf{1.2} & 0.18 & 2.9 & 0.055 \\
    \bottomrule
  \end{tabular}
  }
  \vspace{-1mm}
  \caption{\textbf{Evaluation.} We evaluate our model without Gaussian smoothing (\textit{w/o smoothing}) and compare with the results from the main paper. Note that the success rate and trajectory targets reached (TTR) are not affected as they are evaluated in the physics simulation.}
  \vspace{-3mm}
  \label{tab:no_smoothing}
\end{table*}

%% file: sec/tables/tab_trajectory_per_object.tex
\begin{table}
  \centering
  \resizebox{0.5\linewidth}{!}{%
  \begin{tabular}{@{}lcc@{}}
    \toprule
    Object & Success $\uparrow$  & TTR $\uparrow$\\ \midrule\midrule
    apple & 0.95 & 0.91\\\midrule
    binoculars & 0.54 & 0.83\\\midrule
    camera & 0.89 & 0.85\\\midrule
    mug & 0.64 & 0.74\\\midrule
    toothpaste & 0.94 & 0.94\\
    \bottomrule
  \end{tabular}
  }
  \vspace{-2mm}
  \caption{\textbf{Success And Trajectory Imitation Evaluation.} We evaluate the average success rate and the ratio of trajectory targets reached (TTR) of our method on the GOAL test set. The objects and trajectories are unseen during training.}
  \label{tab:trajectory_per_object}
\end{table}

%% file: sec_supp/X_additional_exp.tex
\section{Additional Experiments}
We provide a more detailed evaluation of two experiments. First, we report the success rate and the trajectory targets reached (TTR) metrics per object of the test set. The results are shown in \cref{tab:trajectory_per_object}. We find that the unseen objects with the most complex shapes, binoculars and mug, have the lowest success rates with 0.54 and 0.64, respectively. A better representation of the object shapes may alleviate such issues in the future. 
Furthermore, we report the metrics without applying Gaussian smoothing to our method in \cref{tab:no_smoothing} (\textit{w/o smoothing}). We find that it helps to improve the ground distance (GD) metric in both the approaching and the manipulation phase. In the approaching phase, it shows less interpenetration. In the manipulation phase, foot skating is reduced when applying smoothing. Moreover, we find the qualitative results to be more visually appealing with smoothing. 

\section{Limitations}
We extend the discussion about limitations of our work and potential future directions from \secref{sec:conclusion}. We consider a unified body shape in our work. Exploring how to vary body shapes is a relatively under-researched problem in physics-based character control and more research is required \cite{christen2023bodyshapes}. Moreover, we use decimation to approximate the object mesh and body shape in order to make the physics simulation sufficiently fast for training. This leads to small interpenetration when converting back to the SMPL-X parametric space. As physics simulations develop, training with more high-resolution meshes will also become feasible. Lastly, our policy struggles with large objects, where the hands have to be fully stretched to grasp. Creating a framework for physics-based two-handed grasping, such as \cite{zhang2023artigrasp}, but for full-body characters may help to overcome such edge cases.

%% file: sec_supp/X_future_work.tex
\section{Ethics Statement}
Our work is in the realm of generating realistic and natural human motion data in simulations. This has future implications in areas such as AR/VR, human-computer interaction (HCI), and robotics. Therefore, one has to be careful in the utilization of such data. While the protection of user data is not a direct concern, since the data we generate is purely synthetic, the downstream use of the data has to be carefully considered. For example, while the generated data may serve in the training of service robots for hospitals or elderly care, it may just as well be used to train military robots. Moreover, being able to generate realistic virtual motions could be misused for generating deep-fakes when combined with realistic rendering techniques. 
While we don't have direct control over the explicit use cases of our technology, we believe discussing potential misuses of the technologies are important. Furthermore, we hope that openly sharing this research, the code and its technical details contributes to understanding the technology and enable access to as many users as possible.